# Double A3C: Deep Reinforcement Learning on OpenAI Gym Games

Yangxin Zhong[1]
Stanford University
Computer Science Department
`yangxin@stanford.edu`

Jiajie He[1]
Stanford University
Mechanical Engineering Department
`jiajie@stanford.edu`

Lingjie Kong[1]
Stanford University
Stanford Center for Professional Development
`ljkong@stanford.edu`

## Abstract

*Reinforcement Learning (RL) is an area of machine learning figuring out how agents take actions in an unknown environment to maximize its rewards. Unlike classical Markov Decision Process (MDP) in which agent has full knowledge of its state, rewards, and transitional probability, reinforcement learning utilizes exploration and exploitation for the model uncertainty. Under the condition that the model usually has a large state space, a neural network (NN) can be used to correlate its input state to its output actions to maximize the agent's rewards. However, building and training an efficient neural network is challenging. Inspired by Double Q-learning and Asynchronous Advantage Actor-Critic (A3C) algorithm, we will propose and implement an improved version of Double A3C algorithm which utilizing the strength of both algorithms to play OpenAI Gym Atari 2600 games to beat its benchmarks for our project.*

## 1. Introduction

Reinforcement Learning (RL) is inspired by behaviorist psychology regarding taking the best actions to optimize agent's reward at a specific state. There have been studies in many disciplines such as control theory, information theory, statistics, and so on.

Classical decision-making problem was formed as a Markov Decision Process (MDP) where people need to have full knowledge of the environment and carefully model its state reward, transitional reward, as well as transitional probability. Due to this limitation, reinforcement learning with Q learning was developed to let agent explore to find possible optimal solution and exploit to optimize the good solutions found up to now.

Under the condition that correlating the large input state space to agent action is not accomplished through look up table like MDP, neural network is used to capture the non-linear relationship between input and output. During the training, forward and backward propagation will be used to train the weight at each layer. With fully trained model, it will be used to inference based on the current state input, what will be the optimal action to take in order to maximize its rewards.

However, building and training an efficient neural network brings challenge in the deep learning perspective. First, deep learning algorithm needs a large amount of data. In particular, RL must be learn from sparse and noisy data collect by the agent which might cause instability. Besides, the reward is usually delayed. To let the reward pass back to prize the initial actions, it requires an efficient method to train the network. Second, neural network model assumes a fix underlying distribution. However, as the agent interact with the real environment, the underlying distribution might not be fixed. Therefore, RL also required a better neural network model which better capture a better model.

This paper will first present Convolutional Neural Networks (CNN) to explain how features will be extracted from each frame of the game. Then, this paper will review classical DQN algorithm to train neural networks with stochastic gradient decent (SGD) with forward and backward propagation. Then, this paper will explain how double DQN will outperform the classical DQN which enables more efficient training. Next, this paper will explain how dueling DQN as well as A3C is a better network structure. Finally, this paper will present Double A3C which utilize the strength from both double QND as well as A3C.

We compare our result on three Atari games: Pong, Breakout, and Ice Hockey.

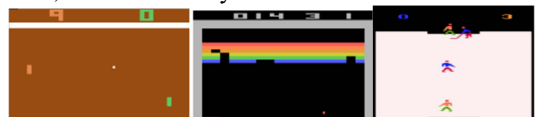

*Figure 1(left to right) Pong, Breakout, Ice Hockey*

## 2. Related Work

Figuring out how to control agent from high-dimensional inputs like vision input is one of the biggest challenges of reinforcement learning. Most successfully RL model before is based on carefully selected feature with linear combined values. Obviously, the quality of the selected feature representation will largely influence the performance.

---

[1]All authors contributed equally to this work.

With the fast development of computer vision, it leads to some breakthroughs on how to extract the feature representation more efficiently by using more efficient models [1]. All these methods utilize ideas of neural network structures such as Convolutional Neural Networks, (CNN), Recurrent Neural Networks (RNN), Multilayer Perception, Boltzmann Machine Graphic Model, and so on.

Besides the challenge from the input feature representation, reinforcement learning presents other challenges. First, traditional machine learning requires a large amount of carefully labelled data, and reinforcement learning algorithm must learn from scalar rewards which is most of the time noisy and delayed from the current state. Second, unlike most supervised learning algorithm which assume the independence of samples, RL's sample are highly correlated.

Q-Learning algorithm [2] with stochastic gradient descent is often used to train reinforcement learning model. In Q-Learning algorithm, we need to store and update a Q value estimate Q(*s, a*) for each (*s, a*) pair, where Q(*s, a*) is the expected utility or value of taking action *a* in state *s* and then following the optimal policy afterwards. However, if we have a large state or action space, it will be expensive to store Q values for all (*s, a*) pairs. One of the common solutions to this issue is to use function approximation, where we extract features $\theta$ from (*s, a*) and define a function $f(\theta)$ to approximate Q(*s, a*). Then optimizing the estimation of Q values turns into optimizing the parameters in $f(\theta)$.

Deep Reinforcement Learning [3] uses a deep neural network, which is called Deep Q-Network (DQN), as the approximate function $f(\theta)$ of Q values. Research showed that the agents trained by DQN can achieve high performances in playing Atari 2600 games in most cases. Further studies of Double DQN [4] and Dueling DQN [5] proposed methods to improve the convergence speed and final performance of DQN. All the DQN models mentioned above can be trained with GPU at a high speed.

Recently, asynchronous method has been proposed to apply to the Deep Reinforcement Learning [6]. The study showed that their best algorithm, Asynchronous Advantage Actor-Critic (A3C), can be trained 2x faster than DQN even if it uses a multi-core CPU instead of GPU. Moreover, agents trained by A3C can achieve higher performances in most of the Atari 2600 games than DQN models.

convolutional neural networks including convolutional layers, maxpooling layers, activation functions, and fully connected layers. The general structure of a convolutional neural network refer to AlexNet [1] as below.

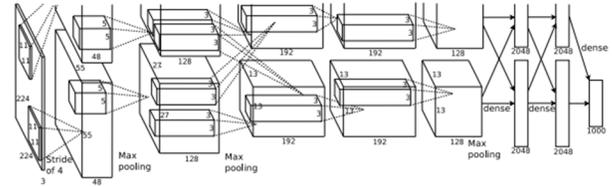

*Figure 2 Convolutional Neural Networks*

Classical DQN will pass casted convolutional layer values into several fully connected layers and eventually generate outputs which has the same dimension of all possible actions. The action which has the higher value will be the optimal action in that specific state.

DQN is usually trained with experience replay which fully utilizing the limited amount of data we have for RL. Basically, the agent will store its previous transition and sample to update its Q value. DQN will be trained by using SGD to minimize the L2 loss between the current Q value and the discounted next Q value by taking the best action plus reward. The overall algorithm [3] is as below.

```
Algorithm 1 Deep Q-learning with Experience Replay
Initialize replay memory D to capacity N
Initialize action-value function Q with random weights
for episode = 1, M do
    Initialise sequence s_1 = {x_1} and preprocessed sequenced φ_1 = φ(s_1)
    for t = 1, T do
        With probability ε select a random action a_t
        otherwise select a_t = max_a Q*(φ(s_t), a; θ)
        Execute action a_t in emulator and observe reward r_t and image x_{t+1}
        Set s_{t+1} = s_t, a_t, x_{t+1} and preprocess φ_{t+1} = φ(s_{t+1})
        Store transition (φ_t, a_t, r_t, φ_{t+1}) in D
        Sample random minibatch of transitions (φ_j, a_j, r_j, φ_{j+1}) from D
        Set y_j = { r_j                              for terminal φ_{j+1}
                  { r_j + γ max_{a'} Q(φ_{j+1}, a'; θ)   for non-terminal φ_{j+1}
        Perform a gradient descent step on (y_j - Q(φ_j, a_j; θ))^2 according to equation 3
    end for
end for
```

*Figure 3 DQN*

However, classical DQN algorithm showed above may suffer from overestimation, and this problem can be improved by utilizing ideas behind the Double Q-learning algorithm. In Double Q-learning, two value $Q^A$ and $Q^B$ functions are learned, with one to determine the greedy policy and the other to determine its value [7]. To be more specifically, double Q-learning algorithm picks $Q^A$ or $Q^B$ randomly to be updated on each step. If $Q^A$ is picked to be updated, then $Q^B$ is used for the value of the next state, shown as follow. Action selections are ε-greedy with respect to the sum of Q1 and Q2.

## 3. Approach

First, the input to the network is each frame of the game in one episode. Each frame was past into convolutional neural network. The general structure of

Algorithm 1 Double Q-learning
1: Initialize $Q^A, Q^B, s$
2: repeat
3:   Choose $a$, based on $Q^A(s, \cdot)$ and $Q^B(s, \cdot)$, observe $r, s'$
4:   Choose (e.g. random) either UPDATE(A) or UPDATE(B)
5:   if UPDATE(A) then
6:     Define $a^* = \arg\max_a Q^A(s', a)$
7:     $Q^A(s, a) \leftarrow Q^A(s, a) + \alpha(s, a)\left(r + \gamma Q^B(s', a^*) - Q^A(s, a)\right)$
8:   else if UPDATE(B) then
9:     Define $b^* = \arg\max_a Q^B(s', a)$
10:     $Q^B(s, a) \leftarrow Q^B(s, a) + \alpha(s, a)(r + \gamma Q^A(s', b^*) - Q^B(s, a))$
11:   end if
12:   $s \leftarrow s'$
13: until end

*Figure 4 Double Q Learning*

Another improvement to classical DQN is the Dueling Network Architecture [5]. Compared to the single stream Q-network, the dueling network has two streams to separately estimate the state value function and the state-dependent action advantage function. It is demonstrated that the dueling architecture can more quickly identify the correct action during policy evaluation as redundant or similar actions are added to the learning problem. The reason why Dueling Network works better is that it separates the value function from its action advantage function and allows the network to capture each individual one better.

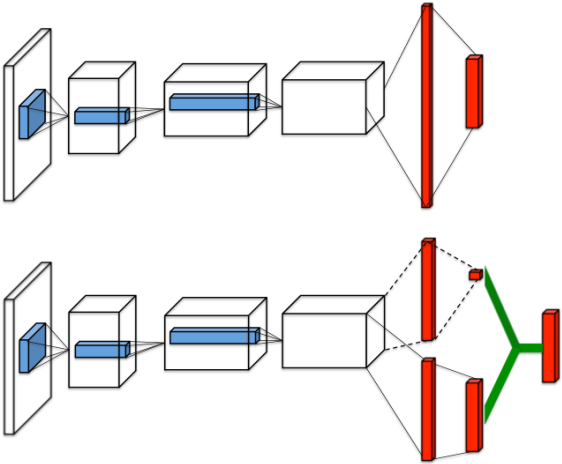

*Figure 5 Single stream Q-network (top) and dueling Q-network (bottom)*

Different from previous method, dueling Q-network defines the network by its state's value as well as its advantage function for a specific action at a state where $\alpha$ and $\beta$ are the parameters of the two streams of fully-connected layers.

$$Q(s, a; \theta, \alpha, \beta) = V(s; \theta, \beta) + A(s, a; \theta, \alpha)$$

The algorithm which has demonstrated the best performances in Atari 2600 domains so far is asynchronous advantage actor-critic (A3C) algorithm. A3C is a multi-threaded asynchronous variant of advantage actor-critic algorithm, where the actor aims at improving the current policy and the critic evaluates the current policy. Like the dueling network architecture, A3C also implements a network which contains two streams to separately update the parameters of the policy and parameters of the value function. The algorithm of A3C is as below [6].

Algorithm S3 Asynchronous advantage actor-critic - pseudocode for each actor-learner thread.
// Assume global shared parameter vectors $\theta$ and $\theta_v$ and global shared counter $T = 0$
// Assume thread-specific parameter vectors $\theta'$ and $\theta'_v$
Initialize thread step counter $t \leftarrow 1$
repeat
  Reset gradients: $d\theta \leftarrow 0$ and $d\theta_v \leftarrow 0$.
  Synchronize thread-specific parameters $\theta' = \theta$ and $\theta'_v = \theta_v$
  $t_{start} = t$
  Get state $s_t$
  repeat
    Perform $a_t$ according to policy $\pi(a_t|s_t; \theta')$
    Receive reward $r_t$ and new state $s_{t+1}$
    $t \leftarrow t + 1$
    $T \leftarrow T + 1$
  until terminal $s_t$ or $t - t_{start} == t_{max}$
  $R = \begin{cases} 0 & \text{for terminal } s_t \\ V(s_t, \theta'_v) & \text{for non-terminal } s_t\text{// Bootstrap from last state} \end{cases}$
  for $i \in \{t-1, \ldots, t_{start}\}$ do
    $R \leftarrow r_i + \gamma R$
    Accumulate gradients wrt $\theta'$: $d\theta \leftarrow d\theta + \nabla_{\theta'} \log \pi(a_i|s_i; \theta')(R - V(s_i; \theta'_v))$
    Accumulate gradients wrt $\theta'_v$: $d\theta_v \leftarrow d\theta_v + \partial (R - V(s_i; \theta'_v))^2 / \partial \theta'_v$
  end for
  Perform asynchronous update of $\theta$ using $d\theta$ and of $\theta_v$ using $d\theta_v$.
until $T > T_{max}$

*Figure 6 A3C*

Our approach will be based on the Double Q learning [7] and state-of-the-art A3C algorithm [6]. The key technique in Double Q learning is to train 2 action-value functions independently. We believe the same technique can also be applied to A3C algorithm. To be more concrete, we add another set of value parameter and randomly pick from one set to update both $\theta$ and $\theta_v$. We hope that by having two value parameters inspired by double Q learning, it can break out the correlation in each sequence of sampled data which leads to faster convergence. We call this method double A3C. There will also be another variant of A3C algorithm introduced in this paper with fewer shared parameters than double A3C, and thus it is called less shared (LS) double A3C.

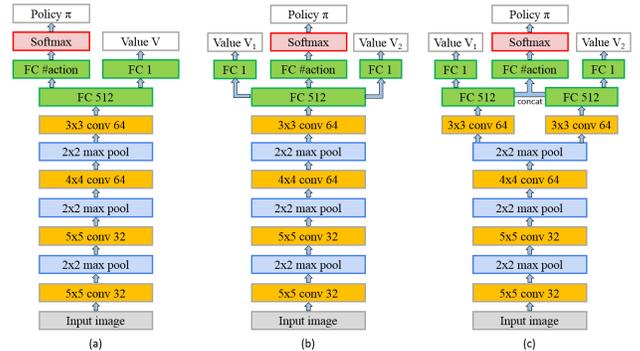

*Figure 7 Network architecture of (a) vanilla A3C, (b) double A3C, (c) less shared (LS) double A3C*

The A3C network architectures that we use are shown in Figure 7. The input we use are 84 x 84 rescaled RGB images of 4 consecutive frames (input size: 84 x 84 x 12). In the network of vanilla A3C, we first apply four convolutional layers along with three max-pooling layers to the input. Details of kernel size and output channel number of each convolutional layer can be found in the figure. On top of the fourth convolutional layer, we apply a fully connected layer with output dimension 512. Then we use this output as the final hidden feature vector $h$. On top of $h$, we apply a fully connected layer with output dimension 1 to get one of the outputs, value estimate V of input/current state. And we also apply a fully connected layer with output dimension $n$, the number of actions, and then a softmax layer to get another output, policy $\pi$ (action probability) of input/current state.

For double A3C network, its outputs are one policy $\pi$ and two different value estimates $V_1$ and $V_2$. The update equation of double A3C is similar to vanilla A3C shown in Figure 6. But in each update step, we randomly pick $V_1$ or $V_2$ to update with equal probability. Suppose we pick $V_1$ to update at step $i$. Then the initialization of R at the last step $t$ will be

$$R \leftarrow \begin{cases} 0 & \text{for terminal } s_t \\ V_2(s_t, \theta'_{v_2}) & \text{for non-terminal } s_t \end{cases}$$

And we also use the following equation to update R down to step $i$

$$R \leftarrow r_i + \gamma R$$

Then we use the following equation to update $d\theta$ and $d\theta_{v_1}$

$$d\theta \leftarrow d\theta + \nabla_{\theta'} \log \pi(a_i|s_i; \theta') \left(R - V_1(s_i; \theta'_{v_1})\right)$$

$$d\theta_{v_1} \leftarrow d\theta_{v_1} + \partial \left(R - V_1(s_i; \theta'_{v_1})\right)^2 / \partial \theta'_{v_1}$$

And if we pick $V_2$ to update in some other steps, we can switch $(V_1, V_2)$, $(\theta_{v_1}, \theta_{v_2})$ and $(\theta'_{v_1}, \theta'_{v_2})$ in the update equations above to get the correct ones.

In the A3C network shown in Figure 7 (a), we can find that $\theta$ and $\theta_v$ share some parameters. The parameters from the first convolutional layer up to the last but two fully connected layers are shared. Only the last fully connected layer (i.e. output layer) are different. Actually, we can see the shared part as a feature extraction network. Then on top of its extracted feature, we can apply different fully connected layer to get policy and value estimate. In deep learning, it's common to use the same extracted feature with several additional fully connected layers to get different outputs. The theory is that the extracted high-level feature can be used to accomplish many different tasks.

For the double A3C network in Figure 7 (b), we use the same idea to share parameters in $\theta$, $\theta_{v_1}$ and $\theta_{v_2}$. That is, we see the network up to the last but two fully connected layers as the feature extraction network and share all their parameters. Then we use different fully connected layers to get the policy $\pi$ and two different value estimates $V_1$ and $V_2$ for current state. The intuitive of using two value estimates is to break the correlation of value estimate of consecutive (probably similar) states. Therefore, one value estimate may help another to get out of local optima. Note that we need two value estimates but only one policy. It's easy to prove that selecting action randomly from one of two policies (two action distributions) is equivalent to selecting action from one policy (a new action distribution).

We also try to share less parameters in double A3C network as shown in Figure 7 (c). We call this method less shared (LS) double A3C. More concretely, we only share parameters up to the third convolutional layer. Then we use two sets of parameters in all other layers. To get the output policy $\pi$, we concatenate the output vectors from two last but two fully connected layers, and then apply an additional fully connected layer and softmax to it. We wonder if less shared parameters can further help to break the correlation in value estimate and thus yields a better performance or a faster convergence.

## 4. Experiment

We train and evaluate our approach using the environment of OpenAI Gym Atari 2600 games. Three games are chosen: Breakout, Ice Hockey, Pong. We implement and compare the average performance of agents trained by vanilla A3C, double A3C, less shared (LS) double A3C, and DQN to measure success. Moreover, we analyze whether our proposed methods will benefit or harm the convergence speed of A3C algorithm.

We set up our reinforcement learning neural network structure as well as OpenAI gym environment based on an existed A3C implementation utilizing tensorpack [8].

A3C, double A3C, and LS double A3C all have 4 convolutional, 3 maxpooling layers and 2 fully connected layers as shown in the previous section to predict value and action benefit individually. Relu is used as the activation function and Adam optimizer is used to minimize the objective loss function. The learning rate is set to 0.001.

Unlike DQN which only uses one specific episode per training step, A3C is concurrently running 3 episodes parallel in multiple thread for reducing the training time. Moreover, unlike DQN which utilizes experience replay consuming lots of memory space to cache all previous state, action, reward tuples for later replay, A3C is using multithreads which does not need to cache previous values.

We also utilize Google Cloud Computing resource to speed up our training. We used Tesla K80 GPU which significantly increase the speed and reduce the time towards convergence. The result we obtained are shown as below.

Our code and implementation can be found under our GitHub repository [9].

Figure 8 shows how the average total reward evolves during training on three games when using vanilla A3C, double A3C, LS double A3C and DQN methods. In Breakout game, three different A3C methods yield dramatically higher mean scores than DQN within the same amount of time of training. In Ice Hockey game, all A3C methods still lead to relatively better mean scores than DQN. In Pong game, all four methods lead to similar results.

Figure 9 compares data efficiency among three different A3C methods. For all three A3C methods, each epoch consists of 6000 steps. It takes around 18 minutes of training per epoch for vanilla A3C and double A3C, and around 20 minutes of training per epoch for LS double A3C.

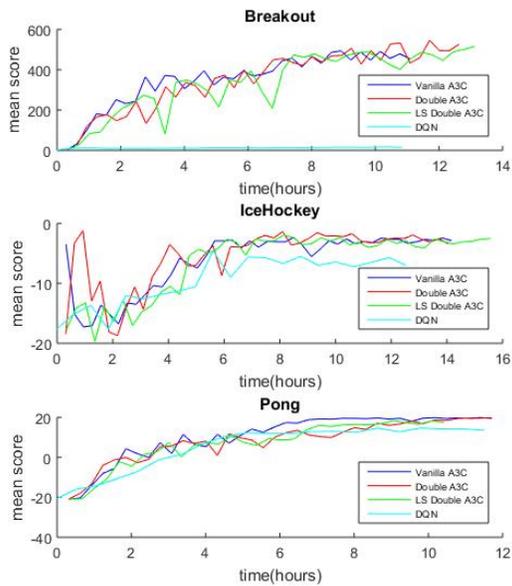

*Figure 8 Average performance and training speed comparison of three different A3C methods and DQN on three Atari games. The x-axis shows the time in hours. The y-axis shows the average score.*

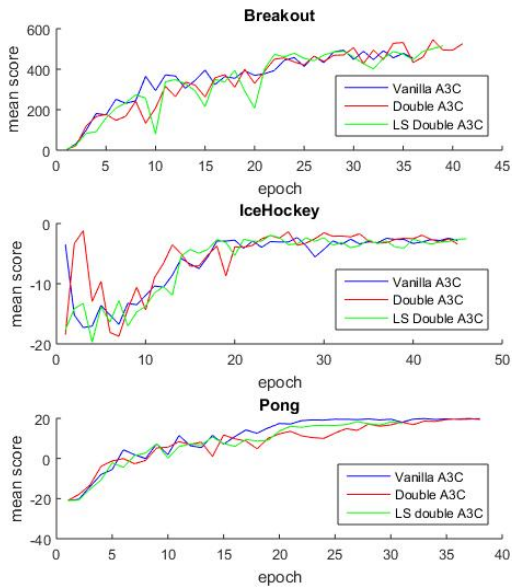

*Figure 9 Data efficiency comparison of three different A3C methods on three Atari games. The x-axis shows the total number of training epochs where an epoch corresponds to 6000 steps. The y-axis shows the average score.*

Table 1 shows the comparison between memory usage and GPU usage between running different algorithms.

*Table 1 Hardware Usage Comparison*

|  | DQN | A3C | Double A3C | LS Double A3C |
|---|---|---|---|---|
| Memory Usage | 8.5G | 1.3G | 1.3G | 1.5G |
| GPU Usage | 23% | 72% | 75% | 86% |

## 5. Discussion

Overall, regarding all three games, no matter what kind of neural network structure we use, Breakout and Pong converge faster compared to Ice Hockey through around 50 epochs with 6000 steps each. Breakout and Pong also takes less time to converge as well. Therefore, there are some levels of complexities that all three types of neural networks (vanilla A3C, double A3C, and LS double A3C) cannot capture in Ice Hockey.

For each game specifically, in Breakout, all A3C structures (vanilla A3C, double A3C, LS double A3C) outperform DQN. Therefore, A3C structure have some advantages over DQN network. First, A3C has multi-threaded asynchronous updates. This allows agents running in parallel to explore different parts of the environment. By running different exploration policies in multiple thread, the update will be less correlated which causes both faster convergence and better results. Second, A3C also utilizes advantage actor-critic structure, instead of using a single output $Q(s,a)$ corresponding to each action at a given state. Advantage actor-critic structure pay attention to value and action benefit individually. Combining both advantages, A3C outperforms DQN.

However, double A3C and LS double A3C have the same performance comparing final score and time of convergence, and they both do not outperform vanilla A3C: Our original hope is that by introducing a second value in the network it will cause updates to be even less correlated to result in better performance and fast convergence time. However, under the condition that A3C structure by utilizing multi-thread update as well as actor-critic structure has already broken the potential correlation in update, double and LS double A3C cannot further enhance the performance. Also, for the update at each epoch, LS double A3C tends to be nosier compared to double A3C. The reason for this phenomenon is that LS double A3C has different last convolutional layers as well as different fully connected layers, while A3C only has different last fully connector layer. Therefore, LS A3C structure might cause two branches predicting two values varying more from each other and cause noisy update.

For Pong, we can see that both DQN and A3C have the same performance. The reason why A3C does not outperform vanilla A3C is that Pong is a simple game without too much interaction when the ball is moving from one side to another. Therefore, its data does not have much correlation from one step to another, so we do not need to rely on A3C structure to break potential correlation. Also, considering its value might be highly related to its action benefit, actor-critic structure will not help to enhance the performance either.

Ice hockey is the most complicated game among all three and no matter what network structure we use, we do not see obvious convergence over time as well as good performance. Besides the fact that Ice Hockey is a complicated game, it is also possible that, even with A3C structure, it still cannot break the correlation during update. Therefore, more studies are needed to analyze why Ice Hockey does not perform well by using A3C structure.

Besides the performance and time of convergence differences among all A3C structures and DQN, we also analyze the memory usage as well as GPU usage. Under the condition that DQN relies on experience replay and previous state, action, reward tuples cached for later use, DQN has higher memory usage compared with all other A3C structures. Regarding GPU, because LS double shared A3C has the most complicated network structure, double A3C is the second, and vanilla A3C is the simplest network structure, during training LS double A3C has the highest GPU usage, double A3C is the second, and vanilla A3C is the third. Under the condition that training relies on the forward and backward propagation, the more complicated the network, the more GPU resource training will require. And also, A3C requires much more GPU resources than DQN. This is because in A3C, we need to run predictors on multi-threads for asynchronous parameters update, which will utilize the GPU more.

## 6. Future Work

There are two major things that we want to cover in our future work.

First, we would like to try other potential neural network structures for gym environment. For example, we have only introduced double-value-and-one-policy method for our double A3C, and we might also want to try double-value-and-double-policy A3C and compare its performance with the current ones. Under the condition that we still have limited computing resource, we cap our convolutional layer up to 3 layers maximum. We also like to try more convolutional layers in the future work to see whether it can capture more features in its state space to enhance the performance for games like Ice Hockey. Besides, we would like to add batch normalization layers after each convolutional layers, regularization, different activation functions, and so on.

Second, we want to analyze why by using the same network structure the results vary from one game environment to another, and we want to utilize some computer vision techniques such as saliency map to see what feature at each state regarding each game contributes most to the decision making. We hope such analytical tool can help us understand each environment better and can enable us to design better neural network structure for each specific environment.

## 7. Conclusion

Through trying different neural network structures such as DQN, A3C, double A3C, and LS double A3C, we realize the neural network is powerful enough to solve reinforcement learning problem to achieve decent performance after hours of training.

However, it is still hard to generalize certain neural network structure to perform well in all gym environments. In the future, we hope to utilize more analytical tools to understand why such network structure does not perform well in certain environment for designing network with even better performance and faster time of convergence.